\documentclass[conference]{IEEEtran}
\IEEEoverridecommandlockouts

\usepackage{cite}
\usepackage{amsmath,amssymb,amsfonts}
\usepackage{algorithmic}
\usepackage{graphicx}
\usepackage{textcomp}
\usepackage[utf8]{inputenc}
\usepackage[table]{xcolor} 
\usepackage{booktabs}
\usepackage{pifont}
\usepackage{multirow}
\usepackage[caption=false]{subfig}
\usepackage{bm}
\usepackage{enumerate}
\usepackage{enumitem}

\definecolor{lightgreen}{RGB}{235,250,235}
\definecolor{medgreen}{RGB}{205,235,205}
\definecolor{deepgreen}{RGB}{170,215,170}

\definecolor{lightyellow}{RGB}{255,255,225}
\definecolor{medyellow}{RGB}{255,248,190}
\definecolor{deepyellow}{RGB}{255,238,150}

\definecolor{lightviolet}{RGB}{245,235,255}
\definecolor{medviolet}{RGB}{225,210,250}
\definecolor{deepviolet}{RGB}{200,180,240}

\definecolor{lightblue}{RGB}{235,245,255}
\definecolor{medblue}{RGB}{205,225,245}
\definecolor{deepblue}{RGB}{170,200,235}

\definecolor{ltgray}{RGB}{245,245,245}
\definecolor{lightteal}{RGB}{225,245,245}
\definecolor{medteal}{RGB}{195,230,230}
\definecolor{deepteal}{RGB}{165,210,210}

\definecolor{deeporange}{RGB}{255, 140, 0}      
\definecolor{medorange}{RGB}{255, 165, 0}       
\definecolor{lightorange}{RGB}{255, 218, 185}   

\def\BibTeX{{\rm B\kern-.05em{\sc i\kern-.025em b}\kern-.08em
    T\kern-.1667em\lower.7ex\hbox{E}\kern-.125emX}}

\begin{document}

\title{\textbf{\texttt{GoCoMA}}: Hyperbolic Multimodal Representation Fusion for Large Language Model- Generated Code Attribution}


\author{
\IEEEauthorblockN{
    Nitin Choudhury\IEEEauthorrefmark{1},
    Bikrant Bikram Pratap Maurya\IEEEauthorrefmark{1},
    Bhavinkumar Vinodbhai Kuwar\IEEEauthorrefmark{1},
    Arun Balaji Buduru
}

\IEEEauthorblockA{
    \IEEEauthorrefmark{1}These authors contributed equally to this work.} 
    {Correspondence: nitinc@iiitd.ac.in}
}

\maketitle

\begin{abstract}

\noindent Large Language Models (LLMs) trained on massive code corpora are now increasingly capable of generating code that is hard to distinguish from human-written code. This raises practical concerns, including security vulnerabilities and licensing ambiguity, and also motivates a forensic question: `Who (or which LLM) wrote this piece of code?' We present \textbf{\texttt{GoCoMA}}, a multimodal framework that models an extrinsic hierarchy between (i) code stylometry, capturing higher-level structural and stylistic signatures, and (ii) image representations of \texttt{binary pre-executable artifacts} (BPEA), capturing lower-level, execution-oriented byte semantics shaped by compilation and toolchains. \textbf{\texttt{GoCoMA}} projects modality embeddings into a hyperbolic Poincar\'e ball, fuses them via a geodesic similarity-based cross-modal attention (GCSA) fusion mechanism, and back-projects the fused representation to Euclidean space for final LLM-source attribution. Experiments on two open-source benchmarks (CoDET-M4 and LLMAuthorBench) show that \textbf{\texttt{GoCoMA}} consistently outperforms unimodal and Euclidean multimodal baselines under identical evaluation protocols.

\end{abstract}


\section{Introduction and Related Work}
\label{sec:intro}

\noindent Large language models (LLMs) are now being extensively used in the software engineering field, from writing code sections and debugging to deployment. This significantly accelerates the software development process, but also raises concerns about its security, licensing, and potential for plagiarism. Recent research also shows evidence for LLM hallucinating against packages~\cite{spracklen2025we}. Considering a scenario where a specific LLM is used for code generation and is later found to produce vulnerable code or be brittle against such package hallucination or generate license-related issues, provenance becomes necessary to retrospectively identify and trace the affected code sections to their originating model. LLMs, such as ChatGPT, Gemini, Claude, Copilot, etc., trained on large code corpora, generate code that is nearly indistinguishable from human-written code~\cite{zeng2025analyst}. These factors combined motivate an urgent need for robust code authorship attribution (CAA) in software forensics~\cite{an2017stack, liu2023reassessing, Zhang2024license, gurioli:hal-04845581}.

\noindent Studies have been exploring CAA for a long time, spanning from human authorship attribution to LLM authorship attribution. Research explored various aspects of features, including naming conventions, stylometry, structural motifs from abstract syntax trees (ASTs), data structures, algorithmic choices, and commenting styles, for CAA~\cite{caliskan2018when, kalgutkar2020code, bogomolov2021authorship, cambronero2024reducing}. However, relying only on source code features can miss deeper behavioral and structural signals. This led to research examining program behavior-specific features while execution~\cite{behavior2016alberto, rosenblum2011wrote}. Based on these feature aspects, researchers explored various statistical and deep learning modeling approaches, which includes, n-gram statistics~\cite{song2022binmlm, paek2024detection}; structural models built from Abstract Syntax Trees, path representations, or token bigrams~\cite{caliskan2015anonymizing, caliskan2015coding, alon2019code2vec}; analyses of runtime or behavioural traces~\cite{wang2018integration}; and deep-learning systems~\cite{alvarez2025clave, alsulami2017source, abuhamad2021large} that learn subtle links between code semantics and author style. In contrast, recent studies show that deep representations can jointly preserve stylometric, structural, and semantic cues, yielding substantially higher attribution accuracy than earlier handcrafted or shallow approaches~\cite{bitton2025detecting, bisztray2025know, gurioli:hal-04845581}. Furthermore, recent research on CAA has been shifting towards watermarking approaches~\cite{kim2025marking, li2025codemark}. LLM watermarking embeds a detectable statistical signature during generation; however, in real-world use, it often becomes unreliable because it only works when the model/provider actually applies the watermarking schemes~\cite{kirchenbauer2023watermark, rastogi2024revisiting, chen2024mark}. Despite extensive exploration, watermarking approaches are not applicable to scenarios where codes generated prior to watermarking, as well as deep representation learning approaches, fail under multilingual setups. \textit{We hypothesize that this gap in deep representation learning is due to unexplored low-level signals derived from binary pre-executable artifacts (BPEAs), and how such byte-level representations relate to higher-level stylistic cues in source code}.

\noindent Addressing these gaps, in this work, we propose \textbf{\texttt{GoCoMA}} (\underline{G}e\underline{o}desic similarity-based \underline{C}r\underline{o}ss-\underline{M}odal \underline{A}ttention), the first framework to \textit{jointly analyze source code and its binary pre-executable artifact (BPEA) as an image, as two complementary multimedia signals for CAA.} We hypothesize that \textit{combining code with BPEA images yields highly discriminative signals for CAA: code PLMs encode high-level lexical, syntactic, and semantic cues, while vision PTMs capture low-level structural and optimization patterns, forming an extrinsic hierarchy. We further posit that Euclidean fusion cannot preserve these hierarchical relations and cross-modal dependencies}, motivating a geometry-aware hyperbolic strategy. \textbf{\texttt{GoCoMA}} encodes both modalities, projects their Euclidean embeddings into the Poincar\'e ball, fuses them with \underline{G}eodesi\underline{C} \underline{S}imilarity–based cross-modal \underline{A}ttention (GCSA), and maps the fused representation back to Euclidean space for classification.

\noindent To summarize, the contribution of the work is as follows--
\begin{itemize}
    \item We introduce a multimodal CAA perspective that jointly models source code representations and vision-based representations derived from transformed binary pre-executable artifacts (BPEA), and we study the resulting process-induced (extrinsic) hierarchy across abstraction levels.
    \item We propose \textbf{\texttt{GoCoMA}}, a geometry-aware hyperbolic multimodal framework that exploits this hierarchy via a geodesic similarity-based cross-modal attention (GCSA) fusion mechanism in the Poincar\'e ball.
    \item We demonstrate improvements over strong unimodal and multimodal Euclidean and hyperbolic baselines, as well as recent LLM-based attribution baselines, on two public benchmarks under identical evaluation protocols.
\end{itemize}

\section{Preliminaries}

\noindent In this section, we discuss the preliminaries of CAA, including BPEA image transformation, the pre-trained models considered for both code and image modalities, and some basics of hyperbolic space representation.

\subsection{Binary Pre-Executable Artifacts (BPEA)}

\noindent Source code written in different programming languages is compiled into their respective intermediate forms, such as class, object, and bytecode files, collectively referred to as binary pre-executable artifacts (BPEA). These raw bytes contain lower, optimization-level information of the source code.

\subsection{Pre-Trained Models (PTMs)}

\noindent \paragraph{Code Pre-trained Language Models (Code PLMs)} 
We utilize four code PLMs--CodeT5+~\cite{wang2023codet5+}, Qwen2.5-Coder~\cite{hui2024qwen2}, UniXcoder~\cite{guo2022unixcoder}, and CodeBERT~\cite{feng2020codebert}--to extract source-level representations (2304, 2048, 768, and 768 dimensions). These models are objectivewise differently trained: CodeT5+ utilizes denoising/span-corruption with code-text alignment, CodeBERT employs MLM and replaced-token detection, UniXcoder employs unified code-text pretraining for structure-aware embeddings, and Qwen2.5-Coder employs causal autoregressive generation for code reasoning. In contrast, vision PTMs over BPEA images capture low-level compilation footprints (e.g., byte-layout and optimization patterns) that are largely inaccessible from source code alone.

\noindent \paragraph{Vision PTMs} 
For image-based BPEA representations, we use four vision PTMs--ConvNeXT-Base~\cite{liu2022convnet}, EfficientNetV2-M~\cite{tan2021efficientnetv2}, ViT-Base (Patch16, 224)~\cite{dosovitskiy2021an}, and MaxViT-T~\cite{tu2022maxvit}--pretrained on large-scale datasets (e.g., ImageNet-21k) but differing in inductive bias. ConvNeXT emphasizes fine-grained local textures, capturing contiguous byte/segment patterns in BPEA images (1024-dimensions). EfficientNetV2-M encodes mid-level hierarchical layouts and clustered regions via scaled CNN representations (1280-dimensions). ViT-Base models long-range spatial dependencies across binary sections using self-attention over $16\times16$ patches, capturing global layout variations (768-dimensions). MaxViT-T combines convolution with multi-axis attention to integrate multi-scale cues from local patterns to overall organization (512-dimensions). We use frozen embeddings (no PTM fine-tuning), and these complementary biases provide a richer BPEA modality for fusion with code PLM representations in \texttt{GoCoMA}.

\subsection{Hyperbolic Fusion}
\noindent Hyperbolic fusion refers to combining modality embeddings after mapping them into a negatively curved hyperbolic space, where distances and aggregation preserve hierarchical structure more effectively than Euclidean operations~\cite{nickel2017poincare, ganea2018hyperbolic, phukan2025hyfuse}. In Euclidean space, standard fusion methods (e.g., concatenation, dot-product attention) can flatten coarse-to-fine relations and suffer from crowding when the underlying signals exhibit tree-like growth, often requiring higher dimensionality to maintain separation. In contrast, hyperbolic spaces (e.g., the Poincar\'e ball or Lorentz model) provide exponential volume growth, allowing fine-grained concepts to spread out while keeping higher-level abstractions closer, which makes hierarchy-aware alignment and aggregation more faithful. In this work, we perform fusion in the Poincar\'e ball by projecting modality embeddings into hyperbolic space, computing cross-modal interactions using hyperbolic geometry, and aggregating them with curvature-aware operations before mapping the fused representation back to Euclidean space for downstream classification.

\section{Dataset}
\noindent We evaluate \texttt{GoCoMA} on two open-source benchmarks for LLM authorship attribution: CoDET-M4~\cite{orel-etal-2025-codet} and LLM-AuthorBench~\cite{bisztray2025know}. CoDET-M4 provides a large-scale, multi-language, multi-domain setting with both human and LLM code, while LLM-AuthorBench targets fine-grained multi-class attribution across closely related, modern code LLMs in a controlled single-language setup.

\subsection{CoDET-M4}
\noindent CoDET-M4 contains 499,467 code samples spanning three programming languages (C++, Java, Python) and three primary sources/domains (LeetCode, CodeForces, GitHub), enabling evaluation under heterogeneous coding styles and contexts. The dataset is nearly balanced, with 252,886 human-written samples and 246,581 LLM-generated samples. LLM code is generated using five models: four locally runnable open-weight code LLMs—CodeLlama-7B, Llama3.1-8B, CodeQwen-1.5-7B, and Nxcode-orpo-7B—and the proprietary GPT-4o as a strong reference generator.


\subsection{LLM-AuthorBench}
\noindent LLM-AuthorBench consists of 32,000 compilable C programs generated from 4,000 unique programming tasks, where each task is implemented by eight state-of-the-art LLMs, yielding a balanced 8-way attribution setting. The generators include GPT-4.1, GPT-4o, GPT-4o-mini, DeepSeek-v3, Qwen2.5-72B, Llama 3.3-70B, Claude-3.5-Haiku, and Gemini-2.5-Flash. The dataset is curated through a process of duplicate removal and compilation.

\section{Methodology}
\noindent In this section, we describe the methodology for the experimentation, including BPEA to image transformation for the considered datasets, followed by the modeling of unimodal, multimodal, and LLM baselines. Finally, we discuss the modeling of \texttt{\textbf{GoCoMA}}. 

\subsection{BPEA to Image Transformation}
\label{subsec:BPEA}

\noindent To construct the visual modality, each source program is first converted into its BPEA form through language-specific compilation and then transformed into a structured image. This BPEA image transformation is inspired by previous literature on malware research, where byte semantics are captured via image format~\cite{freitas2022malnet}, and it also enables pretrained vision models to capture spatial and structural regularities intrinsic to compiled code. 

\noindent Source files are compiled into language-specific pre-executable artifacts: C/C++ code is compiled with \texttt{gcc/clang} into \texttt{.o} object files, Java with \texttt{javac} into \texttt{.class} files (zipped if multiple), and Python with \texttt{py\_compile} into \texttt{.pyc} bytecode. If compilation fails, we fall back to a normalized binary representation by stripping comments/strings/formatting noise and saving it as a binary file. Each artifact is then read as a continuous byte stream and reshaped into a fixed-width RGB image. Consecutive bytes are grouped into triplets $(b_i, b_{i+1}, b_{i+2})$ that form pixel values, arranged row-wise into a 256-pixel-wide grid. Incomplete rows are zero-padded to preserve spatial consistency. The resulting \texttt{.png} images encode both local textures (e.g., opcode-level patterns) and broader layout features that correspond to the structural properties of the compiled binaries.

\subsection{Downstream Baseline Modeling}
\noindent To downstream the PLMs and PTMs, we employ a convolutional neural network (CNN)-based classifier. The embeddings extracted from each model are passed through two 1D convolutional neural network (CNN) layers with 64 and 128 filters (kernel size = 3), followed by max pooling. The resulting feature maps are flattened and fed into a softmax layer to obtain class probabilities. For fusion baselines, we explore both Euclidean and hyperbolic representations: in Euclidean space, features are fused via concatenation and cross-modal attention, while in hyperbolic space, fusion is performed using M\"obius addition.

\noindent \textbf{LLM Baselines.} To benchmark our framework against state-of-the-art Large Language Models (LLMs), we incorporated ChatGPT-5.1, DeepSeek-V3.2-Exp, and Gemini 3 Pro as baselines. We evaluated these models in both zero-shot and few-shot settings for multiclass attribution.

\subsection{\textbf{\texttt{GoCoMA}} Framework} 
\begin{figure}
    \centering
    \includegraphics[width=\linewidth]{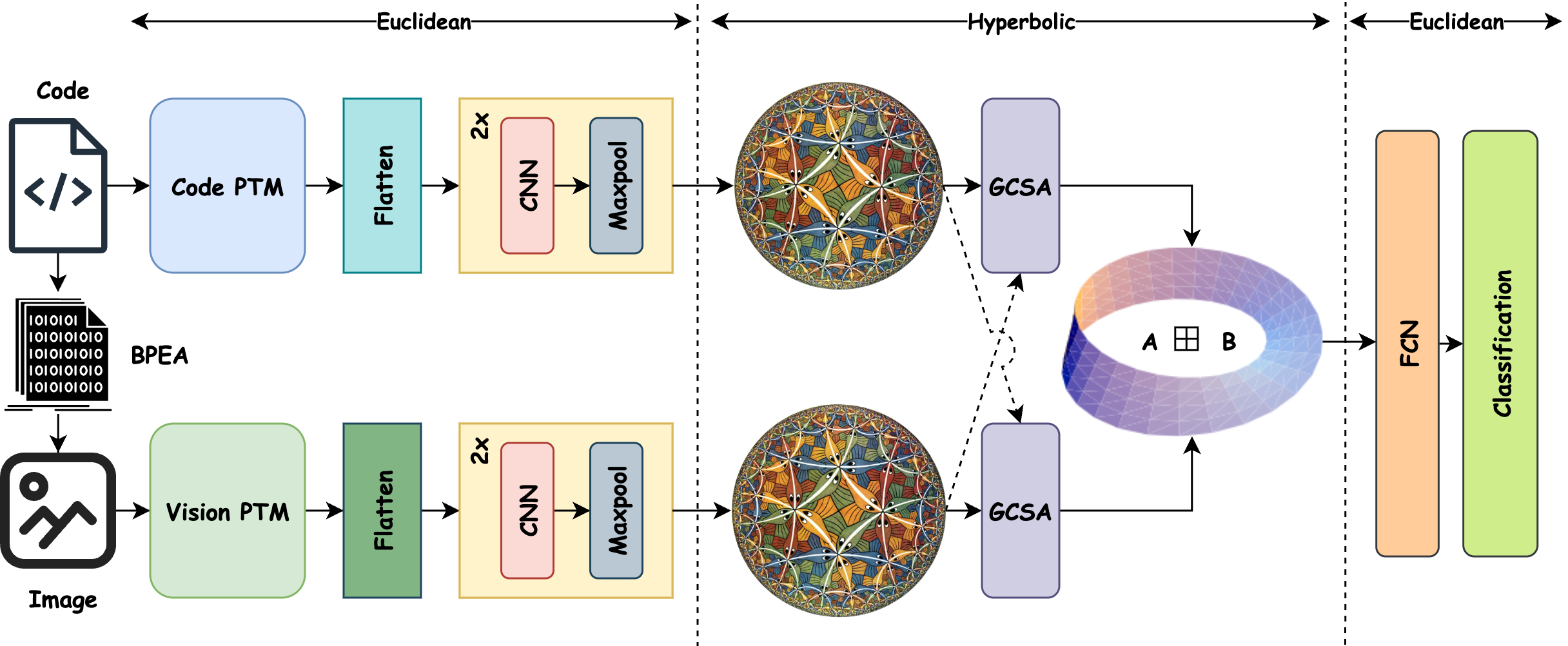}
    \caption{Overview of \texttt{GoCoMA} Architecture}
    \label{fig:gocoma}
\end{figure}

\noindent \textbf{\texttt{GoCoMA}} models the hierarchical relationship and cross-modal dependencies between code and its BPEA image representations within a hyperbolic manifold of constant negative curvature $-c$. \textbf{\texttt{GoCoMA}} comprises three stages. An overview of the \textbf{\texttt{GoCoMA}} framework is shown in Figure~\ref {fig:gocoma}.

\noindent \paragraph{\textbf{Projection from Euclidean to Hyperbolic Space} }
Let $\mathbf{h}^{(E)}_{\text{code}}, \mathbf{h}^{(E)}_{\text{img}} \in \mathbb{R}^n$ denote Euclidean embeddings extracted from the code PLM and vision PTM, respectively. Each embedding is projected to the hyperbolic Poincar\'e ball $\mathbb{D}^n_c = \{\mathbf{x}\in\mathbb{R}^n : c\|\mathbf{x}\|^2 < 1\}$ through the Riemannian exponential map: $\mathbf{h}^{(H)} = \exp_{\mathbf{0}}^c(\mathbf{h}^{(E)})$

\begin{equation}
\mathbf{h}^{(H)} = \tanh\!\left(\sqrt{c}\,\|\mathbf{h}^{(E)}\|\right)\frac{\mathbf{h}^{(E)}}{\sqrt{c}\,\|\mathbf{h}^{(E)}\|}.
\label{eq:exp_map}
\end{equation}

\noindent This transformation embeds Euclidean features into a negatively curved manifold, preserving the hierarchical structure inherent in the modalities.

\paragraph{\textbf{Geodesic Similarity-based Attention (GCSA) Fusion}}
Let $\{\mathbf{h}^{(H)}_{\text{code},i}\}_{i=1}^{T_c}$ and $\{\mathbf{h}^{(H)}_{\text{img},j}\}_{j=1}^{T_v}$ be hyperbolic embeddings in the Poincar\'e ball $\mathbb{D}^d_c$. We define the Query, Key, and Values in this hyperbolic space as: 
\[
\mathbf{Q}_i = \mathcal{L}^c_{W_Q,\mathbf{b}_Q}(\mathbf{h}^{(H)}_{\text{code},i}),
\mathbf{K}_j = \mathcal{L}^c_{W_K,\mathbf{b}_K}(\mathbf{h}^{(H)}_{\text{img},j}), \\
\]
\[
\mathbf{V}_j = \mathcal{L}^c_{W_V,\mathbf{b}_V}(\mathbf{h}^{(H)}_{\text{img},j}).
\]
\noindent These parameters are derived using curvature-consistent M\"obius linear maps: $\mathcal{L}^c_{W,\mathbf{b}}(\mathbf{x}) = (W \otimes_c \mathbf{x}) \oplus_c \mathbf{b}$, $W \otimes_c \mathbf{x} = \exp^c_{\mathbf{0}}\!\big(W\,\log^c_{\mathbf{0}}(\mathbf{x})\big)$

\noindent \textbf{Geodesic Metric.} The geodesic distance on $\mathbb{D}^d_c$ is defined as:
\[
d_c(\mathbf{x},\mathbf{y}) =
\frac{2}{\sqrt{c}}\tanh^{-1}\!\big(\sqrt{c}\,\|-\mathbf{x}\oplus_c\mathbf{y}\|\big),
\]
\noindent where M\"obius addition and M\"obius inner product are defined as:
\[
\mathbf{x}\oplus_c\mathbf{y} =
\frac{(1+2c\langle\mathbf{x},\mathbf{y}\rangle+c\|\mathbf{y}\|^2)\mathbf{x}
+(1-c\|\mathbf{x}\|^2)\mathbf{y}}
{1+2c\langle\mathbf{x},\mathbf{y}\rangle+c^2\|\mathbf{x}\|^2\|\mathbf{y}\|^2},
\]
\[
r\otimes_c\mathbf{x} =
\tanh\!\big(r\,\tanh^{-1}(\sqrt{c}\|\mathbf{x}\|)\big)
\frac{\mathbf{x}}{\sqrt{c}\|\mathbf{x}\|}.
\]

\noindent \textbf{\textbf{\underline{G}eodesi\underline{C} \underline{S}imilarity \underline{A}ttention (GCSA).}}
We define \underline{G}eodesi\underline{c} \underline{S}imilarity (GCS), a distance-based similarity estimator, obtained by applying a cosine transform to the hyperbolic geodesic distance. The similarity between two points is calculated as
\[
\text{GCS}(\mathbf{x},\mathbf{y})
=\cos\!\left(\tfrac{\sqrt{c}}{2}\,d_c(\mathbf{x},\mathbf{y})\right).
\]
Attention coefficients are computed as
\[
s^{(\text{code}\to\text{img})}_{ij} = 
\lambda\,\text{GCS}(\mathbf{Q}_i,\mathbf{K}_j), \quad
s^{(\text{img}\to\text{code})}_{ji} =
\lambda\,\text{GCS}(\mathbf{K}_j,\mathbf{Q}_i),
\]
where $\lambda$ is a learnable temperature.

\noindent \textbf{Curvature-consistent aggregation.}
Each modality aggregates features on the manifold:
\[
\mathbf{z}^{(H)}_{\text{code},i} =
\bigoplus_{j=1}^{T_v}
\big(s^{(\text{code}\to\text{img})}_{ij}\otimes_c\mathbf{V}_j\big), \quad
\]
\[
\mathbf{z}^{(H)}_{\text{img},j} =
\bigoplus_{i=1}^{T_c}
\big(s^{(\text{img}\to\text{code})}_{ji}\otimes_c\mathbf{Q}_i\big).
\]

\noindent \textbf{Joint manifold fusion.} Finally, both modality representations are fused as
\[
\mathbf{z}^{(H)} =
\mathbf{z}^{(H)}_{\text{code}} \oplus_c
\mathbf{z}^{(H)}_{\text{img}},
\]
yielding a unified curvature-aware embedding that preserves inter- and intra-modal dependencies within the hyperbolic space.

\paragraph{\textbf{Back-Projection to Euclidean Space}} The fused hyperbolic vector $\mathbf{z}^{(H)}$ is mapped back to Euclidean space for downstream classification through the logarithmic map: $\mathbf{z}^{(E)} = \log_{\mathbf{0}}^c(\mathbf{z}^{(H)})$

\begin{equation}
\mathbf{z}^{(E)} =
\frac{1}{\sqrt{c}}
\tanh^{-1}\!\left(\sqrt{c}\,\|\mathbf{z}^{(H)}\|\right)
\frac{\mathbf{z}^{(H)}}{\|\mathbf{z}^{(H)}\|}.
\label{eq:log_map}
\end{equation}

\noindent The resulting $\mathbf{z}^{(E)}$ then passed to an FCN followed by a softmax activation for classification.

\section{Experimentation}
\label{sec:experimentation}

\noindent In this section, we outline the experimental protocols, including training and evaluation procedures.

\subsection{Experimentation Protocol}

\paragraph{Data Split}
For \textbf{CoDET-M4}, we utilize the official train-validation-test splits (80-10-10) released with the benchmark to ensure fairness and reproducibility in comparison with existing studies. These partitions are designed to maintain a balanced distribution of human-written and LLM-generated code, while avoiding overlap between the training and evaluation subsets.

\noindent For \textbf{LLMAuthorBench}, in the absence of an official partitioning scheme, we construct an 80-20 stratified split based on the source generator model to preserve class balance. The reported results represent averages from a five-fold cross-validation conducted on the training set, with the reserved 20\% portion used solely for final evaluation.

\noindent For the zero-shot evaluation, we assessed performance based solely on the test set. For the few-shot setting, we incorporated references from the training set for each category before inferring on the test split based on the following prompt.

\begin{quote}
\itshape
``Consider yourself to be a classifier that attributes code snippets to their source LLM. Given a code snippet, decide which model most likely generated it. This is a closed-set, single-label classification problem. You must choose exactly one, the closest label from the following set: [$LLM_1, LLM_2, ..., LLM_n$].''
\end{quote}

\paragraph{Training Procedure and Evaluation Metrics}

\noindent We used the Adam optimizer and categorical cross-entropy to train the model over 25 epochs. With \textit{optuna}, hyperparameters (learning rate, batch size) were adjusted. To prevent overfitting, dropout and early stopping have been incorporated. To promote reproducibility, all experiments were run with fixed random seeds and identical hyperparameter configurations across models. This protocol reduces sampling bias and guards against prompt-level overlap or leakage between training and test partitions. The best hyperparameters identified for optimizing the GoCoMA model are: learning rate of 5e-6, batch size of 16, and dropout of 0.2. Performance is assessed using Accuracy and macro-F1 metrics.

\section{Results and Discussion}
\label{subsec:results}

\begin{table}[h!]
\centering
\scriptsize
\caption{Unified comparison of unimodal/multimodal baselines and LLM prompting baselines on \textbf{LLMAuthorBench} and \textbf{CoDET-M4}. Metrics: Accuracy and macro-F1 (\%). Operators: $+$ (concat), $\otimes$ (Euclidean cross-attn), $\oplus$ (M\"obius), $\boxplus$ (GCSA/\texttt{GoCoMA}); LLM baselines are zero-/few-shot. Abbreviations.: CT5, QWC, UXC, CBE, CNX, EFN, VIT, MXV.}
\label{tab:results}
\begin{tabular}{l|p{1.05cm}p{1.15cm}|p{1.05cm}p{1.15cm}}
\toprule
\multirow{2}{*}{\textbf{Model}} &
\multicolumn{2}{c|}{\textbf{LLMAuthorBench}} &
\multicolumn{2}{c}{\textbf{CoDET-M4}} \\
\cmidrule(lr){2-3} \cmidrule(lr){4-5}
& \textbf{Acc.} & \textbf{macro-F1} & \textbf{Acc.} & \textbf{macro-F1} \\
\midrule
\multicolumn{5}{c}{\textbf{Unimodal - Code (PLMs)}} \\
\midrule
CBE & 93.52 & 85.34 & \cellcolor{deepgreen}83.17 & \cellcolor{deepgreen}75.61 \\
UXC & \cellcolor{medgreen}94.02 & \cellcolor{deepgreen}86.43 & \cellcolor{medgreen}82.70 & \cellcolor{medgreen}74.23 \\
QWC & 91.87 & 83.26 & \cellcolor{lightgreen}81.36 & \cellcolor{lightgreen}73.84 \\
CT5 & \cellcolor{deepgreen}94.13 & \cellcolor{medgreen}86.09 & 79.98 & 71.48 \\
\midrule
\multicolumn{5}{c}{\textbf{Unimodal - Image (PTMs)}} \\
\midrule
CNX & 65.09 & 57.21 & 64.20 & 55.15 \\
EFN & 69.71 & 61.09 & 68.89 & 59.13 \\
VIT & \cellcolor{deepgreen}71.50 & \cellcolor{deepgreen}63.90 & \cellcolor{deepgreen}71.00 & \cellcolor{deepgreen}61.50 \\
MXV & \cellcolor{medgreen}70.28 & \cellcolor{medgreen}61.66 & \cellcolor{medgreen}68.90 & \cellcolor{medgreen}60.37 \\
\midrule
\multicolumn{5}{c}{\textbf{Multi-modality (Model $+$ Model)}} \\
\midrule
CBE $+$ CNX & 94.85 & 86.42 & 84.32 & 75.14 \\
UXC $+$ EFN & 95.18 & 87.45 & 84.65 & 76.29 \\
QWC $+$ MXV & 94.36 & 85.14 & 83.92 & 74.24 \\
CT5 $+$ MXV & 95.48 & 86.19 & 85.77 & 77.15 \\
CT5 $+$ VIT & \cellcolor{lightviolet}95.62 & \cellcolor{lightviolet}87.88 & \cellcolor{lightviolet}86.45 & \cellcolor{lightviolet}78.72 \\
QWC $+$ EFN & 94.63 & 85.36 & 84.22 & 75.12 \\
CBE $+$ VIT & 95.28 & 86.47 & 85.21 & 76.36 \\
\midrule
\multicolumn{5}{c}{\textbf{Multi-modality (Model $\otimes$ Model)}} \\
\midrule
CBE $\otimes$ CNX & \cellcolor{lightviolet}95.58 & 87.02 & 85.62 & 76.93 \\
UXC $\otimes$ EFN & 95.93 & 88.84 & 86.14 & 77.16 \\
QWC $\otimes$ MXV & 95.12 & 86.92 & 85.04 & 75.64 \\
CT5 $\otimes$ MXV & 96.26 & 89.15 & \cellcolor{medviolet}87.73 & \cellcolor{medviolet}78.26 \\
CT5 $\otimes$ VIT & \cellcolor{medviolet}96.35 & \cellcolor{medviolet}90.04 & \cellcolor{medviolet}88.18 & \cellcolor{medviolet}79.57 \\
QWC $\otimes$ EFN & 95.08 & 87.73 & 85.15 & 75.21 \\
CBE $\otimes$ VIT & 95.86 & 88.64 & 86.42 & 77.49 \\
\midrule
\multicolumn{5}{c}{\textbf{Multi-modality (Model $\oplus$ Model, M\"obius Fusion)}} \\
\midrule
CBE $\oplus$ CNX & 96.14 & 88.42 & 86.78 & 77.23 \\
UXC $\oplus$ EFN & 96.31 & 89.73 & 87.12 & 78.04 \\
QWC $\oplus$ MXV & 95.88 & 87.24 & 86.15 & 76.13 \\
CT5 $\oplus$ MXV & 96.97 & 91.06 & \cellcolor{medviolet}88.42 & \cellcolor{medviolet}79.67 \\
CT5 $\oplus$ VIT & \cellcolor{medviolet}97.12 & \cellcolor{medviolet}92.39 & \cellcolor{medviolet}88.84 & \cellcolor{medviolet}80.48 \\
QWC $\oplus$ EFN & 95.98 & 88.17 & 86.35 & 76.23 \\
CBE $\oplus$ VIT & 96.58 & 90.26 & 87.24 & 78.69 \\
\midrule
\multicolumn{5}{c}{\textbf{Multi-modality (Model $\boxplus$ Model, \texttt{GoCoMA})}} \\
\midrule
CBE $\boxplus$ CNX & 96.45 & 88.84 & 86.82 & 77.41 \\
UXC $\boxplus$ EFN & 97.89 & 90.92 & 87.15 & 78.12 \\
QWC $\boxplus$ MXV & 96.12 & 87.95 & 86.47 & 76.61 \\
CT5 $\boxplus$ MXV & \cellcolor{deepviolet}98.56 & 92.85 & \cellcolor{deepviolet}88.24 & \cellcolor{deepviolet}79.72 \\
CT5 $\boxplus$ VIT & \cellcolor{deepviolet}98.92 & \cellcolor{deepviolet}93.82 & \cellcolor{deepviolet}88.67 & \cellcolor{deepviolet}80.97 \\
QWC $\boxplus$ EFN & 96.78 & 88.53 & 86.84 & 76.92 \\
CBE $\boxplus$ VIT & \cellcolor{deepviolet}\textbf{98.81} & \cellcolor{deepviolet}\textbf{94.11} & \cellcolor{deepviolet}\textbf{89.00} & \cellcolor{deepviolet}\textbf{80.17} \\
\midrule
\multicolumn{5}{c}{\textbf{LLM (Zero-Shot)}} \\
\midrule
DeepSeek-V3.2-Exp & 52.41 & 31.87 & 48.60 & 29.00 \\
GPT-5.1           & 61.92 & 49.87 & 56.70 & 45.60 \\
Gemini 3 Pro      & \cellcolor{medblue}66.45 & \cellcolor{medblue}56.38 & \cellcolor{medblue}59.87 & \cellcolor{medblue}50.91 \\
\midrule
\multicolumn{5}{c}{\textbf{LLM (Few-Shot)}} \\
\midrule
DeepSeek-V3.2-Exp & 69.84 & 41.92 & 63.33 & 37.89 \\
Gemini 3 Pro      & \cellcolor{deepblue}83.71 & \cellcolor{deepblue}69.94 & 68.98 & 59.15 \\
GPT-5.1           & 75.22 & 64.18 & \cellcolor{deepblue}75.20 & \cellcolor{deepblue}62.45 \\
\midrule
\multicolumn{5}{c}{SOTA Comparison} \\
\midrule
Bisztra et al.~\cite{bisztray2025know}  & -- & -- & 79.35 & 66.33 \\
Orel et al.~\cite{orel-etal-2025-codet}  & 95.40 & -- & -- & -- \\
\bottomrule
\end{tabular}
\end{table}

\noindent Performance comparison of unimodal and multimodal fusion strategies on \textbf{LLMAuthorBench} and \textbf{CoDET-M4}. Accuracy and macro-F1 are in \%. Here, $+$ denotes concatenation, $\otimes$ Euclidean cross-attention, $\oplus$ hyperbolic fusion via M\"obius addition, and $\boxplus$ hyperbolic fusion via \textbf{\texttt{GoCoMA}.}

\noindent From the empirical evaluation (Table~\ref{tab:results}), it has been observed that the CAA accuracy increases in a clear linear manner once the hierarchy is modeled. Simple concatenation ($+$) yields modest gains by pooling complementary cues. Euclidean cross-attention ($\otimes$) adds a further bump by capturing cross-model dependencies. Moving fusion into hyperbolic space with M\"obius addition ($\oplus$) gives a larger, consistent jump on both benchmarks, indicating that negative curvature better encodes the layered relations among stylometric, syntactic, and semantic signals. Our \textbf{\texttt{GoCoMA}} with geodesic similarity attention ($\boxplus$) extends this advantage: curvature-aware scoring strengthens cross-representation alignment and avoids softmax saturation, producing the strongest results for every PLM-PTM pair.

\noindent Model trends match representational roles. \texttt{CodeBERT} emphasizes surface style, \texttt{UniXcoder} captures structural syntax, and \texttt{CodeT5+} and \texttt{Qwen2.5-Coder} provide deeper semantic and algorithmic cues. On the vision side, \texttt{ViT} aligns best with the compiled-image statistics, consistently outperforming \texttt{ConvNeXt}, \texttt{EfficientNetV2-M}, and \texttt{MaxViT-T} in fusion. The strongest rows pair a high-capacity PLM with \texttt{ViT} under $\boxplus$, with \texttt{CodeBERT}+\texttt{ViT} and \texttt{CodeT5+}+\texttt{ViT} repeatedly topping both accuracy and macro-F1 on LLMAuthorBench and CoDET-M4. Accuracy and macro-F1 move together, indicating balanced gains across authors rather than class skew.

\noindent We further compared \textbf{\texttt{GoCoMA}} against the general-purpose LLM baselines. As shown in Table~\ref{tab:results}, empirical evaluation reveals that \textbf{\texttt{GoCoMA}} significantly surpasses all LLM baselines in both zero-shot and few-shot settings. While few-shot prompting improves LLM performance (e.g., GPT-5.1 improves from 56.70\% to 75.20\%), they still fall short of the geometry-aware fusion approach, which achieves 89.00\% accuracy. This highlights that while general-purpose LLMs are powerful, the specialized hierarchical modeling of \textbf{\texttt{GoCoMA}} is essential for precise authorship attribution.

\noindent Overall, the results validate that the code-binary hierarchy is best captured in hyperbolic space. Geometry-consistent fusion with $\boxplus$ preserves both the depth and interplay of multi-level representations and delivers clear, repeatable lifts over Euclidean concatenation and cross-attention, pushing LLMAuthorBench near the ceiling and yielding strong margins on CoDET-M4.

\noindent \textbf{Ablation Study.} Table~\ref{tab:results} summarizes ablations over modality, fusion, and geometry. Although code PLMs consistently outperform BPEA-image PTMs in an unimodal setting, indicating that source code contains stronger authorship cues, BPEA still provides a meaningful low-level signal, as evidenced by their comparable performance in CoDET-M4. Fusing code with BPEA in Euclidean space outperforms unimodal results, demonstrating that BPEA provides complementary evidence beyond source stylometry. Replacing concatenation with Euclidean cross-attention yields further gains, confirming the value of explicit cross-modal interaction. The largest and most consistent improvements appear when fusion is moved to hyperbolic space: hyperbolic GCSA better aligns byte-level BPEA semantics with high-level code representations, and the full GoCoMA pipeline (hyperbolic GCSA plus Möbius fusion) achieves the best accuracy and macro-F1 overall, highlighting that BPEA provides the extra attribution signal and hyperbolic GCSA is the key module that makes this signal exploitable.

\noindent \textbf{Limitations.} Our evidence comes from two public benchmarks, so generalization to other languages, repositories, and build systems is uncertain. The binary-to-image pathway is sensitive to the toolchain: compilers, optimization flags, linkers, and packaging can change byte layouts and shift the image distribution. We analyze static artifacts only, omitting runtime traces and build metadata that could help separate closely related generators. Robustness is untested: style obfuscation, formatting normalization, semantics-preserving edits, or binary padding may degrade performance.

\section{Conclusion}
\label{sec:conclusion}

\noindent We introduced \textbf{\texttt{GoCoMA}}, a geometry-aware multimodal framework that couples code PLM representations with image features from binary pre-executable artifacts and fuses them in a Poincar\'e ball using GCSA. Across both LLMAuthorBench and CoDET-M4, \textbf{\texttt{GoCoMA}} outperforms unimodal and Euclidean multimodal baselines, supporting the hypothesis that authorship cues form a hierarchy that is preserved more faithfully in hyperbolic space. Practically, the strongest recipe is a high-capacity code PLM paired with ViT, fused via GCSA, which delivers consistent gains in both accuracy and macro-F1.



\bibliographystyle{IEEEbib}
\bibliography{icme2026references}

@inproceedings{spracklen2025we,
  title={We Have a Package for You! A Comprehensive Analysis of Package Hallucinations by Code Generating $\{$LLMs$\}$},
  author={Spracklen  et al., Joseph},
  booktitle={34th USENIX Security Symposium (USENIX Security 25)},
  pages={3687--3706},
  year={2025}
}

@article{zeng2025analyst,
  title={An Analyst-Inspector Framework for Evaluating Reproducibility of LLMs in Data Science},
  author={Zeng et al., Qiuhai},
  journal={arXiv preprint arXiv:2502.16395},
  year={2025}
}

@inproceedings{caliskan2015anonymizing,
  title={De-anonymizing programmers via code stylometry},
  author={Caliskan-Islam et al.},
  booktitle={24th USENIX security symposium (USENIX Security 15)},
  pages={255--270},
  year={2015}
}

@INPROCEEDINGS{behavior2016alberto,
  author={Ferrante et al., Alberto},
  booktitle={2016 11th International Conference on Availability, Reliability and Security (ARES)}, 
  title={Spotting the Malicious Moment: Characterizing Malware Behavior Using Dynamic Features}, 
  year={2016},
  volume={},
  number={},
  pages={372-381},
  doi={10.1109/ARES.2016.70}
}

@inproceedings{rosenblum2011wrote,
  title={Who wrote this code? identifying the authors of program binaries},
  author={Rosenblum, Nathan and Zhu, Xiaojin and Miller, Barton P},
  booktitle={European Symposium on Research in Computer Security},
  pages={172--189},
  year={2011},
  organization={Springer}
}

@inproceedings{wang2018integration,
  title={Integration of static and dynamic code stylometry analysis for programmer de-anonymization},
  author={Wang et al., Ningfei},
  booktitle={Proceedings of the 11th ACM workshop on artificial intelligence and security},
  pages={74--84},
  year={2018}
}

@article{bisztray2025know,
  title={I Know Which LLM Wrote Your Code Last Summer: LLM generated Code Stylometry for Authorship Attribution},
  author={Bisztray et al., Tamas},
  journal={arXiv preprint arXiv:2506.17323},
  year={2025}
}

@article{bitton2025detecting,
  title={Detecting Stylistic Fingerprints of Large Language Models},
  author={Bitton et al., Yehonatan},
  journal={arXiv preprint arXiv:2503.01659},
  year={2025}
}

@article{kim2025marking,
  title={Marking Code Without Breaking It: Code Watermarking for Detecting LLM-Generated Code},
  author={Kim, Jungin and Park, Shinwoo and Han, Yo-Sub},
  journal={arXiv preprint arXiv:2502.18851},
  year={2025}
}

@article{alon2019code2vec,
  title={code2vec: Learning distributed representations of code},
  author={Alon, Uri and Zilberstein, Meital and Levy, Omer and Yahav, Eran},
  journal={Proceedings of the ACM on Programming Languages},
  volume={3},
  number={POPL},
  pages={1--29},
  year={2019},
  publisher={ACM New York, NY, USA}
}

@article{caliskan2015coding,
  title={When coding style survives compilation: De-anonymizing programmers from executable binaries},
  author={Caliskan, Aylin and Yamaguchi, Fabian and Dauber, Edwin and Harang, Richard and Rieck, Konrad and Greenstadt, Rachel and Narayanan, Arvind},
  journal={arXiv preprint arXiv:1512.08546},
  year={2015}
}

@article{abuhamad2021large,
  title={Large-scale and robust code authorship identification with deep feature learning},
  author={Abuhamad, Mohammed and Abuhmed, Tamer and Mohaisen, David and Nyang, Daehun},
  journal={ACM Transactions on Privacy and Security (TOPS)},
  volume={24},
  number={4},
  pages={1--35},
  year={2021},
  publisher={ACM New York, NY, USA}
}

@inproceedings{alsulami2017source,
  title={Source code authorship attribution using long short-term memory based networks},
  author={Alsulami, Bander and Dauber, Edwin and Harang, Richard and Mancoridis, Spiros and Greenstadt, Rachel},
  booktitle={European Symposium on Research in Computer Security},
  pages={65--82},
  year={2017},
  organization={Springer}
}

@article{alvarez2025clave,
  title={CLAVE: A deep learning model for source code authorship verification with contrastive learning and transformer encoders},
  author={{\'A}lvarez-Fidalgo, David and Ortin, Francisco},
  journal={Information Processing \& Management},
  volume={62},
  number={3},
  pages={104005},
  year={2025},
  publisher={Elsevier}
}

@article{wang2023codet5+,
  title={Codet5+: Open code large language models for code understanding and generation},
  author={Wang, Yue and Le, Hung and Gotmare, Akhilesh Deepak and Bui, Nghi DQ and Li, Junnan and Hoi, Steven CH},
  journal={arXiv preprint arXiv:2305.07922},
  year={2023}
}

@article{feng2020codebert,
  title={Codebert: A pre-trained model for programming and natural languages},
  author={Feng, Zhangyin and Guo, Daya and Tang, Duyu and Duan, Nan and Feng, Xiaocheng and Gong, Ming and Shou, Linjun and Qin, Bing and Liu, Ting and Jiang, Daxin and others},
  journal={arXiv preprint arXiv:2002.08155},
  year={2020}
}

@article{hui2024qwen2,
  title={Qwen2. 5-coder technical report},
  author={Hui, Binyuan and Yang, Jian and Cui, Zeyu and Yang, Jiaxi and Liu, Dayiheng and Zhang, Lei and Liu, Tianyu and Zhang, Jiajun and Yu, Bowen and Lu, Keming and others},
  journal={arXiv preprint arXiv:2409.12186},
  year={2024}
}

@article{guo2022unixcoder,
  title={Unixcoder: Unified cross-modal pre-training for code representation},
  author={Guo, Daya and Lu, Shuai and Duan, Nan and Wang, Yanlin and Zhou, Ming and Yin, Jian},
  journal={arXiv preprint arXiv:2203.03850},
  year={2022}
}

@inproceedings{gurioli:hal-04845581,
  TITLE = {{Is This You, LLM? Recognizing AI-written Programs with Multilingual Code Stylometry}},
  AUTHOR = {Gurioli, Andrea and Gabbrielli, Maurizio and Zacchiroli, Stefano},
  URL = {https://hal.science/hal-04845581},
  BOOKTITLE = {{IEEE International Conference on Software Analysis, Evolution and Reengineering, SANER 2025}},
  ADDRESS = {Montréal, Canada},
  YEAR = {2025},
  MONTH = Mar,
  HAL_ID = {hal-04845581},
  HAL_VERSION = {v1},
}

@inproceedings{orel-etal-2025-codet,
    title = "{C}o{D}et-M4: Detecting Machine-Generated Code in Multi-Lingual, Multi-Generator and Multi-Domain Settings",
    author = "Orel et al., Daniil",
    editor = "Che, Wanxiang  and
      Nabende, Joyce  and
      Shutova, Ekaterina  and
      Pilehvar, Mohammad Taher",
    booktitle = "Findings of the Association for Computational Linguistics: ACL 2025",
    month = jul,
    year = "2025",
    address = "Vienna, Austria",
    publisher = "Association for Computational Linguistics",
    url = "https://aclanthology.org/2025.findings-acl.550/",
    doi = "10.18653/v1/2025.findings-acl.550",
    pages = "10570--10593",
    ISBN = "979-8-89176-256-5"
}

@inproceedings{liu2023reassessing,
      title={Reassessing Code Authorship Attribution in the Era of Language Models},
      author={Liu et al., Andy},
      booktitle={32nd USENIX Security Symposium (USENIX Security 23)},
      year={2023},
      isbn={978-1-939133-37-3},
      pages={257--274},
      url={https://www.usenix.org/conference/usenixsecurity23/presentation/liu-andy},
      publisher={USENIX Association}
}

@article{Zhang2024license,
  author    = {Zhang et al., Tianyi},
  title     = {A First Look at License Compliance Capability of LLMs in Code Generation},
  journal   = {arXiv preprint arXiv:2408.02487},
  eprint    = {2408.02487},
  archivePrefix = {arXiv},
  year      = {2024}
}

@inproceedings{an2017stack,
  author    = {An, Le and Mlouki, Ons and Khomh, Foutse and Antoniol, Giuliano},
  title     = {Stack Overflow: A Code Laundering Platform?},
  booktitle = {24th International Conference on Software Analysis, Evolution, and Reengineering (SANER)},
  pages     = {282--292},
  year      = {2017},
  address   = {Klagenfurt, Austria},
  publisher = {IEEE},
  doi       = {10.1109/SANER.2017.7884629}
}

@article{kalgutkar2020code,
  author    = {Vaibhav Kalgutkar and Rupinder Kaur and Hern{\'a}n Gonzalez and Natalia Stakhanova and Anita Matyukhina},
  title     = {Code Authorship Attribution: Methods and Challenges},
  journal   = {ACM Computing Surveys},
  volume    = {52},
  number    = {1},
  pages     = {1--36},
  year      = {2020},
  doi       = {10.1145/3292577}
}

@inproceedings{caliskan2018when,
  author    = {Aylin Caliskan and Fabian Yamaguchi and Engin Dauber and Richard Harang and Konrad Rieck and Rachel Greenstadt and Arvind Narayanan},
  title     = {When Coding Style Survives Compilation: De-anonymizing Programmers from Executable Binaries},
  booktitle = {Proceedings of the Network and Distributed System Security Symposium (NDSS)},
  year      = {2018},
  doi       = {10.14722/ndss.2018.23304}
}

@inproceedings{bogomolov2021authorship,
  author    = {Evgeny Bogomolov and Vladyslav Kovalenko and Yaroslav Rebryk and Alberto Bacchelli and Timofey Bryksin},
  title     = {Authorship Attribution of Source Code: A Language-Agnostic Approach and Applicability in Software Engineering},
  booktitle = {Proceedings of the 29th ACM Joint Meeting on European Software Engineering Conference and Symposium on the Foundations of Software Engineering (ESEC/FSE)},
  pages     = {1183--1195},
  year      = {2021},
  doi       = {10.1145/3468264.3468606}
}

@article{cambronero2024reducing,
  author    = {José Cambronero and Francisco J. Rodríguez and Laura Moreno and Daniel M. German and Premkumar Devanbu},
  title     = {Reducing the Impact of Time Evolution on Source Code Authorship Attribution via Unsupervised Data Augmentation},
  journal   = {Proceedings of the ACM on Software Engineering},
  volume    = {1},
  number    = {FSE},
  pages     = {1--25},
  year      = {2024},
  doi       = {10.1145/3652151}
}

@article{liu2022convnet,
  title     = {A ConvNet for the 2020s},
  author    = {Liu, Zhuang and Mao, Hanzi and Wu, Chao-Yuan and Feichtenhofer, Christoph and Darrell, Trevor and Xie, Saining},
  journal   = {Proceedings of the IEEE/CVF Conference on Computer Vision and Pattern Recognition (CVPR)},
  year      = {2022},
  pages     = {11976--11986}
}

@inproceedings{tan2021efficientnetv2,
  title     = {EfficientNetV2: Smaller Models and Faster Training},
  author    = {Tan, Mingxing and Le, Quoc V.},
  booktitle = {Proceedings of the 38th International Conference on Machine Learning (ICML)},
  year      = {2021},
  pages     = {10096--10106}
}

@inproceedings{dosovitskiy2021an,
  title     = {An Image is Worth 16x16 Words: Transformers for Image Recognition at Scale},
  author    = {Dosovitskiy, Alexey and Beyer, Lucas and Kolesnikov, Alexander and Weissenborn, Dirk and Zhai, Xiaohua and Unterthiner, Thomas and Dehghani, Mostafa and Minderer, Matthias and Heigold, Georg and Gelly, Sylvain and Uszkoreit, Jakob and Houlsby, Neil},
  booktitle = {International Conference on Learning Representations (ICLR)},
  year      = {2021}
}

@inproceedings{tu2022maxvit,
  title     = {MaxViT: Multi-Axis Vision Transformer},
  author    = {Tu, Zhengzhong and Talebi, Hossein and Zhang, Han and Yang, Feng and Milanfar, Peyman and Bovik, Alan C. and Mittal, Anish},
  booktitle = {Proceedings of the European Conference on Computer Vision (ECCV)},
  year      = {2022},
  pages     = {459--479}
}

@inproceedings{paek2024detection,
  title={Detection of LLM-Generated Java Code Using Discretized Nested Bigrams},
  author={Paek, Timothy and Mohan, Chilukuri},
  booktitle={International Conference on Computational Science and Computational Intelligence},
  pages={118--132},
  year={2024},
  organization={Springer}
}

@inproceedings{song2022binmlm,
  title={BinMLM: Binary authorship verification with flow-aware mixture-of-shared language model},
  author={Song, Qige and Zhang, Yongzheng and Ouyang, Linshu and Chen, Yige},
  booktitle={2022 IEEE International Conference on Software Analysis, Evolution and Reengineering (SANER)},
  pages={1023--1033},
  year={2022},
  organization={IEEE}
}

@article{li2025codemark,
  title={CodeMark: Contextual and Natural Watermarking for Tracing Code Snippet Provenance},
  author={Li, Wei and Yang, Borui and Sun, Yujie and Chen, Suyu and Chen, Yuting and Xiang, Liyao},
  journal={IEEE Transactions on Dependable and Secure Computing},
  year={2025},
  publisher={IEEE}
}

@inproceedings{kirchenbauer2023watermark,
  title={A watermark for large language models},
  author={Kirchenbauer, John and Geiping, Jonas and Wen, Yuxin and Katz, Jonathan and Miers, Ian and Goldstein, Tom},
  booktitle={International Conference on Machine Learning},
  pages={17061--17084},
  year={2023},
  organization={PMLR}
}

@article{rastogi2024revisiting,
  title={Revisiting the robustness of watermarking to paraphrasing attacks},
  author={Rastogi, Saksham and Pruthi, Danish},
  journal={arXiv preprint arXiv:2411.05277},
  year={2024}
}

@article{chen2024mark,
  title={De-mark: Watermark removal in large language models},
  author={Chen, Ruibo and Wu, Yihan and Guo, Junfeng and Huang, Heng},
  journal={arXiv preprint arXiv:2410.13808},
  year={2024}
}

@inproceedings{freitas2022malnet,
  title={MalNet: A large-scale image database of malicious software},
  author={Freitas, Scott and Duggal, Rahul and Chau, Duen Horng},
  booktitle={Proceedings of the 31st ACM International Conference on Information \& Knowledge Management},
  pages={3948--3952},
  year={2022}
}

@article{nickel2017poincare, 
  title={Poincar{\'e} embeddings for learning hierarchical representations},
  author={Nickel, Maximillian and Kiela, Douwe},
  journal={Advances in neural information processing systems},
  volume={30},
  year={2017}
}

@article{ganea2018hyperbolic, 
  title={Hyperbolic neural networks},
  author={Ganea, Octavian and B{\'e}cigneul, Gary and Hofmann, Thomas},
  journal={Advances in neural information processing systems},
  volume={31},
  year={2018}
}

@article{phukan2025hyfuse,
  title={HYFuse: Aligning Heterogeneous Speech Pre-Trained Representations in Hyperbolic Space for Speech Emotion Recognition},
  author={Phukan, Orchid Chetia and Akhtar, Mohd Mujtaba and Behera, Swarup Ranjan and Reddy, Pailla Balakrishna and Buduru, Arun Balaji and Sharma, Rajesh and others},
  journal={arXiv preprint arXiv:2506.03403},
  year={2025}
}
\end{document}